\documentclass[letterpaper]{article} 
\usepackage{aaai2026}  
\usepackage{times}  
\usepackage{helvet}  
\usepackage{courier}  
\usepackage[hyphens]{url}  
\usepackage{graphicx} 
\usepackage{natbib}  
\usepackage{caption} 
\usepackage{algorithm}
\usepackage{algorithmic}

\usepackage{newfloat}
\usepackage{listings}
\DeclareCaptionStyle{ruled}{labelfont=normalfont,labelsep=colon,strut=off} 
\floatstyle{ruled}
\newfloat{listing}{tb}{lst}{}
\floatname{listing}{Listing}
\usepackage{multirow}
\usepackage{booktabs}
\usepackage{amsmath}
\usepackage{amsfonts}
\usepackage{subcaption}
\usepackage{pifont}
\usepackage{xcolor}
\newcommand{\xmark}{\ding{55}} 

\nocopyright 

\title{Symmetrical Flow Matching: Unified Image Generation, Segmentation, and Classification with Score-Based Generative Models}
\author{
    Francisco Caetano\textsuperscript{\rm 1},
    Christiaan Viviers\textsuperscript{\rm 1},
    Peter H.N. De~With\textsuperscript{\rm 1},
    Fons van der Sommen\textsuperscript{\rm 1}
}
\affiliations{
    \textsuperscript{\rm 1}Eindhoven University of Technology, The Netherlands\\

    f.caetano@tue.nl
}

\begin{document}

\maketitle

\begin{abstract}
Flow Matching has emerged as a powerful framework for learning continuous transformations between distributions, enabling high-fidelity generative modeling. This work introduces Symmetrical Flow Matching~(SymmFlow), a new formulation that unifies semantic segmentation, classification, and image generation within a single model. Using a symmetric learning objective, SymmFlow models forward and reverse transformations jointly, ensuring bi-directional consistency, while preserving sufficient entropy for generative diversity. A new training objective is introduced to explicitly retain semantic information across flows, featuring efficient sampling while preserving semantic structure, allowing for one-step segmentation and classification without iterative refinement. Unlike previous approaches that impose strict one-to-one mapping between masks and images, SymmFlow generalizes to flexible conditioning, supporting both pixel-level and image-level class labels. Experimental results on various benchmarks demonstrate that SymmFlow achieves state-of-the-art performance on semantic image synthesis, obtaining FID scores of 11.9 on CelebAMask-HQ and 7.0 on COCO-Stuff with only 25 inference steps. Additionally, it delivers competitive results on semantic segmentation and shows promising capabilities in classification tasks.
\end{abstract}

\begin{links}
     \link{Code}{https://github.com/caetas/SymmetricFlow}
\end{links}

\section{Introduction}
\label{sec:intro}

Comprehending semantic content is a key challenge in computer vision. Classification~\cite{krizhevsky2012imagenet, he2016deep, dosovitskiy2020image} and segmentation~\cite{ronneberger2015u, xie2021segformer, cheng2022masked} allow models to analyze and structure images, while generative modeling enables the synthesis of new content~\cite{radford2015unsupervised, ho2020denoising, rombach2022high}. Ideally, a unified framework would bridge these tasks, allowing models to both interpret and generate images in a two-way manner. It is conceivable that the ability to accurately comprehend and disentangle visual structures facilitates the generation of more semantically coherent and visually realistic images. Conversely, strong generative capabilities may aid in learning more expressive representations of images, as generating plausible content requires an implicit understanding of object relationships, textures, and context~\cite{he2022masked}. These forward and backward relationships suggest that advances in one direction could naturally benefit the other, which fuels the pursuit of models that integrate both understanding and synthesis within a single cohesive framework.

Most existing vision-only approaches treat these tasks individually. For classification, the extracted features are used in fully connected layers to obtain probability predictions for each class, while for segmentation, the features are used to train decoders for dense predictions. On the other hand, generative models, such as generative adversarial networks~(GANs)~\cite{karras2021alias}, diffusion models~\cite{karras2022elucidating}, score matching models~\cite{song2020score}, and Flow Matching models~\cite{lipman2022flow}, synthesize images from a prior distribution. Although recent work has explored diffusion models for classification~\cite{li2023your} and segmentation~\cite{wu2024medsegdiffv2}, these adaptations introduce significant limitations: classification is slow due to the need for iterative sampling across all possible classes, and segmentation frameworks are restricted to generating masks, lacking the ability to map back to realistic images.

Recent works, such as SemFlow~\cite{wang2024semflow} and DepthFM~\cite{gui2024depthfm}, have sought to unify image generation with semantic segmentation and depth estimation within a single generative framework. However, these models still suffer from the following key limitations: (1)~they do not perform classification, restricting their applicability; (2)~the image quality remains inferior to that of purely generative models; and (3)~the models enforce a strict one-to-one mapping between segmentation or depth masks and images, requiring them to have the same number of channels, which limits their flexibility.

\begin{figure*}[!t]
    \centering
    \includegraphics[width=0.99\linewidth]{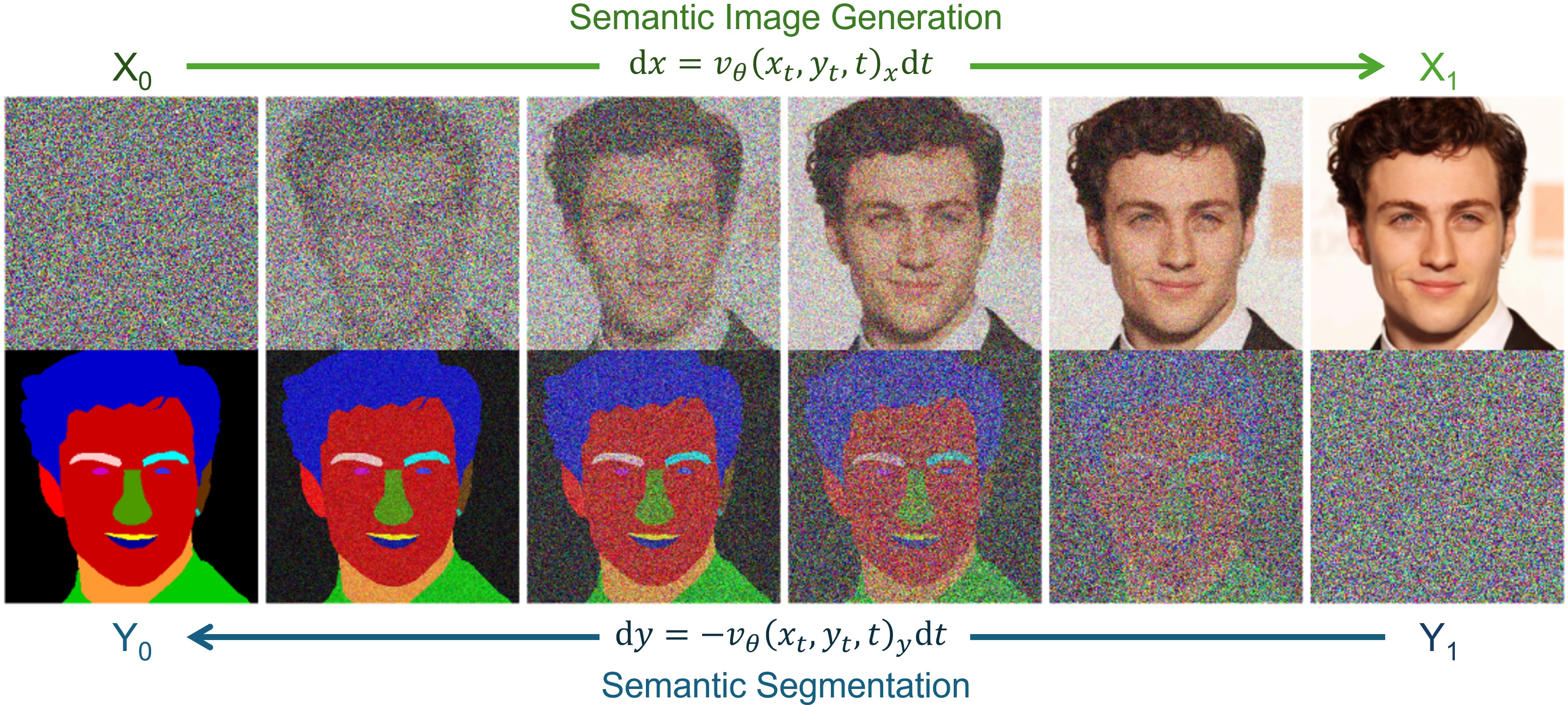}
    \caption{Symmetrical Flow Matching jointly models semantic segmentation and generation as opposing flows. Noise transitions into an image while a label evolves into noise and vice versa. This symmetry maintains entropy for generation while enforcing semantic consistency. Image Y can represent semantic content of any type, from dense masks to global labels, enabling applications like classification and segmentation.}
    \label{fig: symmflow}
\end{figure*}

To address these limitations, we propose Symmetrical Flow Matching~(SymmFlow), a novel Flow Matching training objective that enforces dual symmetrical sampling, enabling segmentation, classification, and image synthesis to be conditioned in a mutual two-sided relationship. This concept leads to the following specific contributions. (a)~SymmFlow unifies segmentation and classification within a single framework, performing both tasks in fewer steps, while retaining the ability to generate high-quality images through Flow Matching. (b)~SymmFlow enhances image synthesis quality over prior methods by leveraging the bi-directionality of Flow Matching. (c)~SymmFlow alleviates the strict one-to-one channel constraint between segmentation masks and images, allowing greater flexibility in conditioning and generalization. We validate SymmFlow on toy problems as well as standard benchmarks for classification, segmentation, and image synthesis, demonstrating its effectiveness as a unified model for both discriminative and generative tasks.

\section{Related Work}
\label{sec:related}

\subsection{Flow Matching}

Flow Matching~(FM)~\cite{lipman2022flow, liu2022flow} is a generative modeling framework that learns a velocity field to transform a source distribution into a target distribution through a continuous flow. By parameterizing this field with a neural network, the model becomes a Neural ODE~\cite{chen2018neural}, allowing efficient sampling via numerical integration. It has advanced state-of-the-art performance in diverse applications, including image~\cite{esser2024scaling} and audio generation~\cite{vyas2023audiobox}, as well as protein modeling~\cite{huguet2024sequence} and robotics~\cite{black2024pi_0}. FM generalizes Continuous Normalizing Flows~(CNFs)~\cite{chen2018neural, grathwohl2018ffjord} by eliminating the need for simulation during training, thereby making it computationally efficient. It also provides a unifying perspective on generative modeling, encompassing diffusion models, which can be seen as a special case where the probability path is defined via stochastic differential equations~\cite{lipman2024flow}. 

\subsection{Generative Classifiers}

Several seminal works~\cite{hinton2007recognize, ranzato2011deep} have emphasized the importance of modeling the data distribution to enhance discriminative feature learning. Early approaches trained deep belief networks~\cite{hinton2006fast} to encode image data as latent representations, which were then used for recognition tasks. More recent advances in generative modeling have demonstrated the ability to learn efficient representations for global prediction tasks~\cite{he2022masked}. In addition, generative models have been shown to improve adversarial robustness and calibration~\cite{huang2020neural}. However, most prior work either jointly trains generative and discriminative models, or fine-tunes generative representations for downstream tasks. Diffusion Classifier~\cite{li2023your} specifically investigates the effectiveness of using diffusion models as image classifiers, albeit with severe inference constraints.

\subsection{Semantic Segmentation}

Semantic segmentation aims to assign a semantic label to each pixel in an image. Conventional approaches rely on discriminative models, combining a strong feature extraction backbone with a task-specific decoder head for mask prediction~\cite{cheng2021per, cheng2022masked, ding2023mevis}. Recent work has explored diffusion models for segmentation~\cite{amit2021segdiff, wang2024explore, gu2024diffusioninst}, typically leveraging them as feature extractors within a discriminative framework. A key motivation behind SemFlow is the interpretation that diffusion models struggle to align their stochastic nature with the deterministic requirements of semantic segmentation. For this reason, SemFlow introduces rectified flows. In contrast, SymmFlow embraces the probabilistic nature of segmentation, accounting for inter-observer variability.

\subsection{Semantic Image Generation}

Semantic image generation, the inverse of semantic segmentation, focuses on generating realistic images from semantic layouts~\cite{li2019diverse, liu2019learning, lv2022semantic}. Existing approaches generally fall into two categories. (1)~GAN-based models~\cite{zhu2017unpaired, isola2017image, wang2018high}, although many of these methods struggle with mode collapse and produce only unimodal outputs. (2)~Diffusion models, which treat semantic generation as a conditional generation task where semantic masks act as control signals~\cite{wang2022semantic}. Some methods further integrate additional conditioning to enhance coherence~\cite{zhang2023adding}. However, these approaches often adopt asymmetric architectures with unidirectional generators, thereby complicating the unification of semantic segmentation and image generation.

\section{Symmetrical Flow Matching}
\label{sec:symmflow}

Symmetrical Flow Matching~(SymmFlow) unifies semantic segmentation and semantic synthesis as opposing flow processes, as illustrated in Figure~\ref{fig: symmflow}. Given a data distribution $X$ (e.g. images) and a semantic representation $Y$ (e.g. masks or class labels), SymmFlow models bi-directional flows between them. The forward process transforms $X$ from noise, while simultaneously evolving $Y$ towards a noise-corrupted state. The reverse process inverts these transitions, allowing for the generation of $Y$ from $X$. Crucially, $Y$ is not restricted to having the same dimensionality as $X$, enabling flexible conditioning, such as global class labels for classification. This symmetrical formulation ensures sufficient entropy for image generation while preserving the semantic structure, making SymmFlow a generalizable framework for both segmentation and synthesis. The training procedure and objective are formalized in Section~\ref{subsec:training}. Techniques for obtaining segmentation and classification predictions using the proposed SymmFlow model are discussed in Section~\ref{subsec:classify&segment}. The importance of label dequantization for stable training is examined in Section~\ref{subsec:dequant}, and the complete framework is validated using a synthetic example presented in Section~\ref{subsec:toy}. 

\subsection{Training Objective}
\label{subsec:training}

\begin{figure}[!ht]
    \centering
    \includegraphics[width=\linewidth]{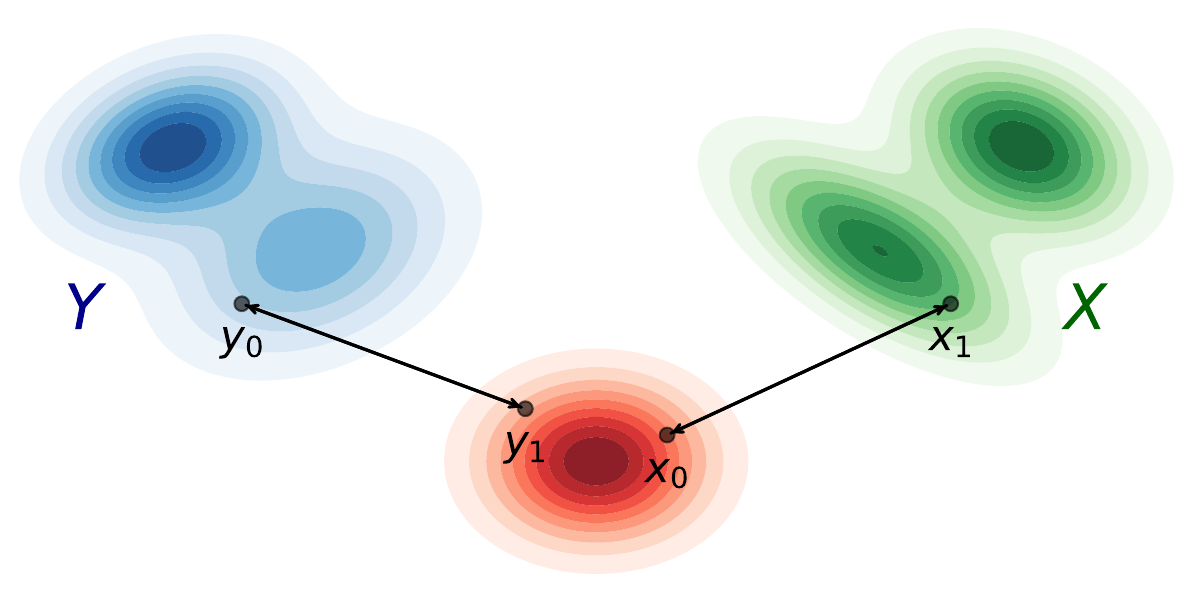}
    \caption{Illustration of the optimal transport between the data distributions X and Y, and the intermediate Gaussian distribution.}
    \label{fig: dist}
\end{figure}

Symmetrical Flow Matching jointly models semantic segmentation and synthesis as opposing flows, enabling bi-directional transformations between images and semantic content. The model learns a velocity field that transports $X$ from noise ($X_0$), while simultaneously evolving $Y$ into noise and vice versa. 
For each sample, a time variable $t$ is extracted from $\mathcal{U}(0,1)$, and the inputs are perturbed via a convex combination with Gaussian noise. As a result, the perturbed samples $x_t$~(forward) and $y_t$~(backward) are specified by  
\begin{equation}
\begin{aligned}
    x_t &= (1 - t) \xi_x + t x, \\
    y_t &= (1 - t) y + t \xi_y, 
\end{aligned}
\end{equation}
where $\xi_x, \xi_y$ are independent noise terms extracted from $\mathcal{N}(0, I)$. The optimal transport velocity fields are given by  
\begin{equation}
\begin{aligned}
    v_x &= x - \xi_x, \\
    v_y &= \xi_y - y,  \\
    v &= (v_x, v_y),
\end{aligned}
\end{equation}
which describe the ideal directions to reverse the perturbation. Figure~\ref{fig: dist} illustrates the optimal transport approach in Symmetrical Flow Matching, showing the transformation between data distributions and the Gaussian intermediary.

The model $v_\theta(x_t, y_t, t)$ is trained to jointly approximate both flows by minimizing the squared error, specified by 
%
\begin{equation}
\mathcal{L} = \mathbb{E}_{x, y, t} \left[ \| v_\theta(x_t, y_t, t) - v \|^2 \right].
\end{equation}

\subsection{Classification and Segmentation}
\label{subsec:classify&segment} 

A common approach to classification using conditional generative models relies on Bayes’ theorem to compute the posterior probability of a class $c$ given an input image $X$. Given a generative model that learns the conditional distribution $p_\theta(x \mid c)$, classification is performed as
\begin{equation}
    p_\theta(c_i \mid x) = \frac{p(c_i) p_\theta(x \mid c_i)}{\sum_j p(c_j) p_\theta(x \mid c_j)}.
\end{equation}
With a uniform prior over classes, i.e.  $p(c_i) = \frac{1}{N}$, the prior terms cancel, simplifying to  
\begin{equation}
    p_\theta(c_i \mid x) \propto p_\theta(x \mid c_i).
\end{equation}
For diffusion models, computing $p_\theta(x \mid c)$ is intractable, so an ELBO approximation is used to estimate the posterior distribution 
\begin{equation}
    p_\theta(c_i \mid x) = \frac{\exp\{-\mathbb{E}_{t,\epsilon}[\,\|\epsilon - \epsilon_\theta(x_t, c_i)\|^2\,]\}}{\sum_j \exp\{-\mathbb{E}_{t,\epsilon}[\,\|\epsilon - \epsilon_\theta(x_t, c_j)\|^2\,]\}}.
\end{equation}  
Monte Carlo sampling approximates the expectation by
\begin{equation}
    \frac{1}{N} \sum_{i=1}^{N} \|\epsilon_i - \epsilon_\theta(\sqrt{\bar{\alpha}_{t_i}} x + \sqrt{1 - \bar{\alpha}_{t_i}} \epsilon_i, c_j)\|^2.
\end{equation} 
Ultimately, this approach extracts a classifier by evaluating the error between noise predictions for each class.

\subsubsection{Proposed Approach} 

In contrast to the conventional generative classifier approach, SymmFlow learns a velocity field that transports an input image toward a noise distribution, and vice versa. Classification is performed by integrating the predicted velocity field in an off-the-shelf Ordinary Differential Equation~(ODE) solver
\begin{equation}
    y_0 = y_1 + \int_1^0 v_\theta(x_t, y_t, t)_y dt.
\end{equation}
This eliminates the need for repeated evaluations across all possible class embeddings, significantly reducing inference time and computational cost. Additionally, the same process can be used to predict segmentation masks. The predicted class is determined as the closest label to the average of the model’s predictions. For segmentation, the class of each pixel is assigned based on the closest predefined class RGB code to the predicted pixel RGB value. This mapping is further explained in Section~\ref{subsec:toy}.

\subsection{Dequantization}\label{subsec:dequant} 

In line with prior work, we dequantize the labels (classification and segmentation) to a continuous distribution. Dequantization is often used in Normalizing Flows to enhance stability in density modeling. Without it, excessively high likelihoods (dirac deltas) are assigned to a few specific values, causing the model to collapse. 
A standard approach to dequantization involves adding finite perturbations to the signal to prevent low-entropy distributions from hindering modeling quality. We adopt a similar strategy by applying controlled noise to the class labels $Y$, ensuring smoother optimization and preventing degenerated solutions. Specifically, given a discrete label $Y$, we define the dequantized representation as  
\begin{equation}  
Y' = Y + \epsilon, \quad \epsilon \sim U(-\beta, +\beta),  
\end{equation}  
where $U(-\beta, +\beta)$ is a uniform noise term ensuring that the semantic label remains well-defined. This adjustment is crucial for maintaining stability in the reverse flow process.  

For models performing classification, we further normalize the label representations to the interval [-1, +1], based on their indices, prior to supplying them to the model. The dequantized values then serve as a continuous mask, providing a structured conditioning mechanism for the model.

\subsection{Toy Example}
\label{subsec:toy}

\begin{figure}[!ht]
    \centering
    \begin{subfigure}[t]{0.49\linewidth}
        \centering
        \includegraphics[width=\linewidth]{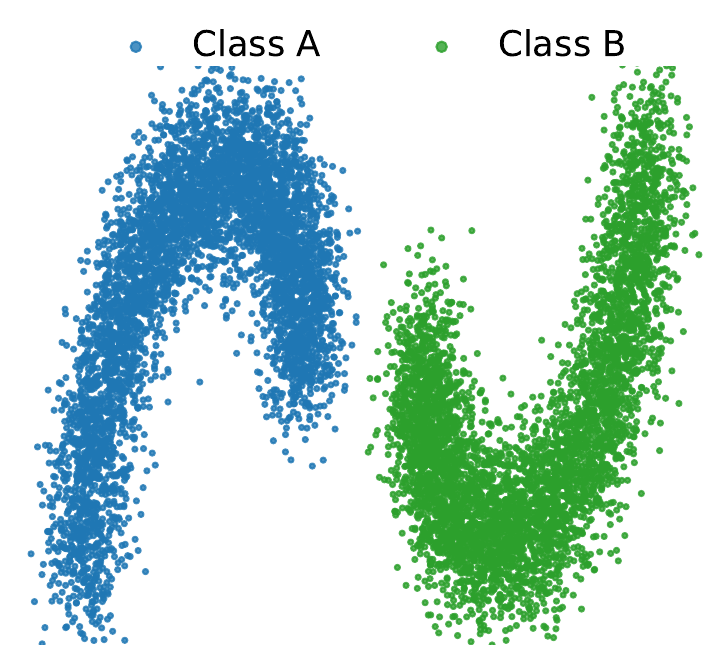}
        \caption{Original spiral dataset.}
        \label{fig:spirals_data}
    \end{subfigure}
    \hfill
    \begin{subfigure}[t]{0.49\linewidth}
        \centering
        \includegraphics[width=\linewidth]{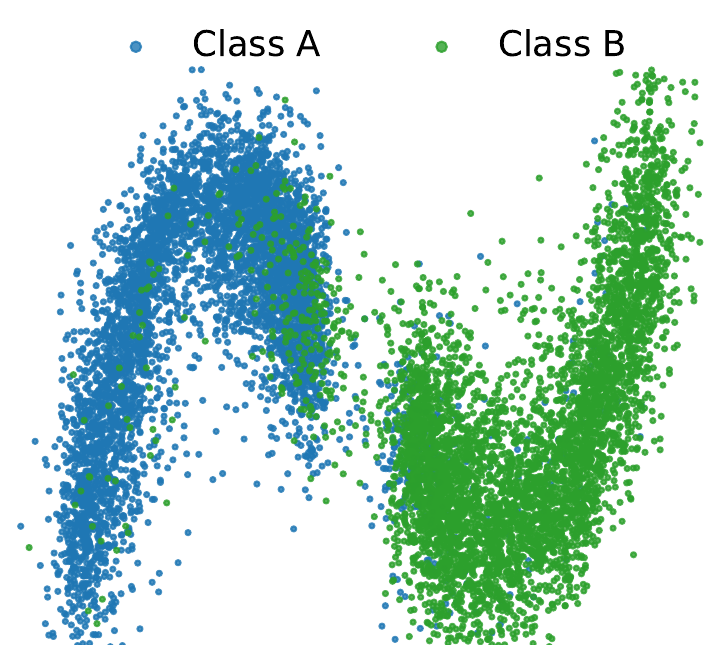}
        \caption{Generated samples.}
        \label{fig:spirals_generated}
    \end{subfigure}
    \caption{Visualization of the spiral dataset and samples generated by our model.}
    \label{fig:spirals}
\end{figure}

To illustrate the principles of Symmetrical Flow Matching, we consider a toy example where two classes form the interleaved, nonlinear structures in Figure~\ref{fig:spirals_data}.

We train a multilayer perceptron~(MLP) to model the joint evolution of data points $X$ and their class representation $Y$ under Symmetrical Flow Matching. The input to the model consists of the point coordinates $X$, a quantized encoding of their class $Y$ - where Class A follows a uniform distribution in [-1.5, -0.5] and Class B in [+0.5, +1.5] — and a time variable $t$ ranging from 0 to 1. The learned flow is then used to sample trajectories via an Euler ODE solver with 20 timesteps from $t=0$ to $t=1$, reconstructing the underlying structure of the data. The resulting distributions, shown in Figure~\ref{fig:spirals_generated}, demonstrate the ability of our framework to model structured transformations while maintaining class separability.

We also model the reverse process, integrating the learned flow backward from $t = 1$ to $t = 0$ to recover class labels. This formulation treats classification as an inverse problem, requiring the model to separate class information as the distribution regresses to its original state. Table~\ref{tab:accuracy_vs_steps} shows that the classification accuracy is highest when using a single integration step. This result could be expected: as $X$ evolves toward a Gaussian distribution, class boundaries blur, making it increasingly difficult for the reverse flow to correctly infer the original labels.

\begin{table}[ht]
    \centering
    {\resizebox{\linewidth}{!}{
    \begin{tabular}{l|c c c c c c}
        \toprule
        \textbf{N\textsubscript{steps}} & 1 & 2 & 5 & 10 & 20 & 50 \\
        \midrule
        \textbf{Acc.~(\%)} & \textbf{100.0} & 92.0 & 87.0 & 83.6 & 82.6 & 82.0 \\
        \bottomrule
    \end{tabular}}}
    \caption{Classification accuracy at different numbers of steps using the reverse process of the model.}
    \label{tab:accuracy_vs_steps}
\end{table}

\section{Experiments}
\label{sec:experiments}
To evaluate the effectiveness of the proposed SymmFlow model as a unified architecture for semantic segmentation/classification and image synthesis, we perform the following experiments.

\begin{table*}[t]
    \centering
    \renewcommand{\arraystretch}{1.3}
    \resizebox{\linewidth}{!}{
    \begin{tabular}{l|l|c|cc|cc}
        \toprule
        \multirow{2}{*}{\textbf{Category}} & \multirow{2}{*}{\textbf{Method}} & \multirow{2}{*}{\textbf{Steps}} & \textbf{SS~(mIoU$\uparrow$)} & \textbf{SS~(mIoU$\uparrow$)} & \textbf{SIS~(FID$\downarrow$~\slash~LPIPS$\uparrow$)} & \textbf{SIS~(FID$\downarrow$~\slash~LPIPS$\uparrow$)} \\ 
        & & & \textbf{CelebAMask-HQ} 
        & \textbf{COCO-stuff} 
        & \textbf{CelebAMask-HQ} 
        & \textbf{COCO-stuff} \\
        \midrule
        \multirow{5}{*}{SS}
        & DML-CSR~\cite{zheng2022decoupled} & 1 & 77.8 & --- & \xmark & \xmark \\
        & SegFace~\cite{narayan2025segface} & 1 & \textbf{81.6} & --- & \xmark & \xmark \\
        & DeeplabV2~\cite{caesar2018coco} & 1 & --- & 33.2 & \xmark & \xmark \\
        & MaskFormer~\cite{cheng2021per} & 1 & --- & 37.1 & \xmark & \xmark \\
        & SegFormer~\cite{xie2021segformer} & 1 & --- & \textbf{46.7} & \xmark & \xmark \\
        \midrule
        \multirow{7}{*}{SIS}
        & pix2pixHD~\cite{wang2018high} & 1 & \xmark & \xmark & 54.7~\slash~0.529 & 111.5~\slash~--- \\
        & SPADE~\cite{park2019semantic} & 1 & \xmark & \xmark & 42.2~\slash~0.487 & 33.9~\slash~--- \\
        & SC-GAN~\cite{wang2021image} & 1 & \xmark & \xmark & 19.2~\slash~0.395 & 18.1~\slash~--- \\
        & BBDM~\cite{li2023bbdm} & 200 & \xmark & \xmark & 21.4~\slash~0.370 & --- \\
        & ControlNet~\cite{zhang2023adding} & 20 & \xmark & \xmark & 24.0~\slash~0.528 & 36.6~\slash~0.671 \\
        & SDM~\cite{wang2022semantic} & 1000 & \xmark & \xmark & 18.8~\slash~0.422 & 15.9~\slash~0.518 \\
        & SCDM~\cite{ko2024stochastic} & 250 & \xmark & \xmark & 17.4~\slash~0.418 & 15.3~\slash~0.519 \\
        & SCP-Diff~\cite{gao2024scp} & 800 & \xmark & \xmark & --- & 11.3~\slash~--- \\
        \midrule
        \multirow{2}{*}{Both}
        & SemFlow~\cite{wang2024semflow} & 25 & \hphantom{*}69.4$^*$ & \hphantom{*}35.7$^*$ & 32.6~\slash~0.393 & \hphantom{*}90.0~\slash~0.685$^*$ \\
        & SymmFlow~(Proposed) & 25 & 69.3 & 39.6 & \textbf{11.9}~\slash~0.464 & \textbf{7.0}~\slash~0.609 \\
        \bottomrule
    \end{tabular}
    }
    \caption{Performance comparison of benchmark solutions on semantic segmentation~(SS) and semantic image synthesis~(SIS) tasks across both COCO-Stuff and CelebAMask-HQ datasets. The number of steps indicates the number of functions applied to obtain the results. Legend: $^*$~Recomputed by us; ---~Not available; \xmark~Not applicable (method cannot perform the task).}
    \label{tab:sys_results}
\end{table*}

\subsection{Datasets}
\emph{Semantic Segmentation and Generation:} The model is evaluated using COCO-Stuff~\cite{caesar2018coco} and CelebAMask-HQ~\cite{lee2020maskgan}, which contain 171 and 19 classes, respectively. Images and semantic masks are resized and cropped into 512$\times$512 pixels. A detailed analysis of the datasets is provided in Appendix~\ref{sec:data_sup}. \emph{Classification:} We evaluate our model on MNIST~\cite{deng2012mnist} and CIFAR-10~\cite{krizhevsky2010cifar} for classification, leveraging these low-resolution datasets to assess fundamental capabilities.

\subsection{Metrics}

\emph{Semantic Segmentation and Generation:} For semantic segmentation, we evaluate with mean intersection over union~(mIoU). For semantic image synthesis, we assess with the Fréchet inception distance~(FID) and learned perceptual image patch similarity~(LPIPS). \emph{Classification:} The model is evaluated using classification accuracy. Additional information is available in Appendix~\ref{sec:metrics_sup}.

\subsection{Qualitative Evaluation}
To complement the quantitative analysis, qualitative results are presented for classification, semantic segmentation, and image generation. For classification, generated samples from MNIST and CIFAR-10 are visualized in Appendix~\ref{sec:results_sup} to evaluate the diversity and fidelity of the synthesized outputs. For semantic segmentation, predicted masks on CelebAMask-HQ and COCO-stuff are compared against ground-truth annotations. For image generation, representative samples synthesized from CelebAMask-HQ and COCO-stuff are provided to illustrate visual quality.

\subsection{Impact of Inference Steps}  

The effect of the number of inference steps is evaluated for both tasks. By varying the number of steps, the impact on classification accuracy and segmentation quality is assessed, providing insight into the trade-off between computational efficiency and performance.

\subsection{Baseline Comparison}

\emph{Semantic Segmentation and Generation:} For semantic segmentation on CelebAMask-HQ, DML-CSR~\cite{zheng2022decoupled} and SegFace~\cite{narayan2025segface} are adopted as baselines. On COCO-Stuff, DeeplabV2~\cite{caesar2018coco}, MaskFormer~\cite{cheng2021per}, and SegFormer~\cite{xie2021segformer} serve as representative methods. For semantic image synthesis, the evaluation includes pix2pixHD~\cite{wang2018high}, SPADE~\cite{park2019semantic}, SC-GAN~\cite{wang2021image}, BBDM~\cite{li2023bbdm}, ControlNet~\cite{zhang2023adding}, SDM~\cite{wang2022semantic}, SCDM~\cite{ko2024stochastic}, and SCP-Diff~\cite{gao2024scp}. SemFlow~\cite{wang2024semflow}, a prior model aiming to unify both segmentation and synthesis, is also evaluated. Since official checkpoints and semantic segmentation evaluation scripts for SemFlow were not publicly available, the model was retrained and results were recomputed. The scripts are included in SymmFlow's repository for reproducibility. \emph{Classification:} For the classification task on CIFAR-10, the proposed model is compared to the Diffusion Classifier~\cite{li2023your}. Additionally, the quality of the generated images is assessed by benchmarking against a standard FM setup.

\subsection{Implementation Details}

For pixel-level implementations~(e.g., classification), the U-Net architecture introduced in Guided Diffusion~\cite{dhariwal2021diffusion} is employed. For latent-space models, the pre-trained VAE from Stable Diffusion~\cite{rombach2022high} is used for image encoding and decoding, alongside the U-Net backbone from Stable Diffusion 2.1. To ensure compatibility with SymmFlow, the number of input channels in the first layer and output channels in the final layer of the U-Net are doubled. Further details on hyperparameters and computational resources are provided in Appendix~\ref{sec:implementation_sup}.

\begin{figure*}[!ht]
    \centering
    \includegraphics[width=\linewidth]{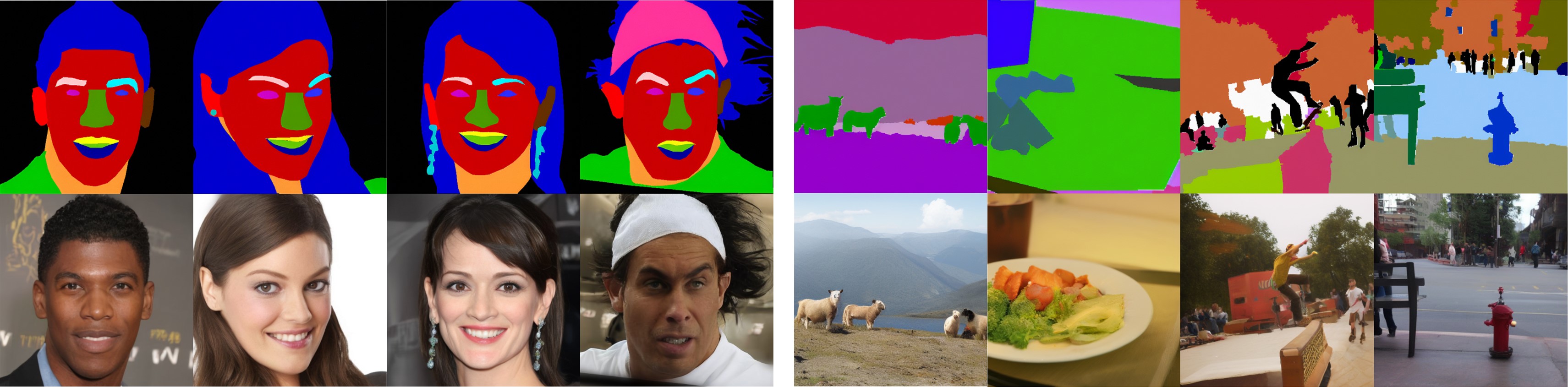}
    \caption{Non-curated samples generated by the model trained on CelebAMask-HQ~(left) and COCO-stuff~(right). The top row shows the semantic mask used to condition the model. The bottom row shows the samples after 25 integration steps with the Euler ODE solver.}
    \label{fig:celeba_samples}
\end{figure*}

\section{Results \& Discussion}
\label{sec:results}

\subsection{Semantic Segmentation and Generation}

Table~\ref{tab:sys_results} reports semantic segmentation and image synthesis results across benchmarks. SymmFlow consistently outperforms prior models in the synthesis task, achieving the lowest FID scores across datasets. The visualizations in Figure~\ref{fig:celeba_samples} confirm that the model produces mask-consistent high-fidelity samples on CelebAMask-HQ and COCO-Stuff, capturing structural details with strong adherence to the conditioning masks. Although LPIPS is commonly used to assess perceptual diversity, it can be misleading in isolation, as higher LPIPS values may result from poor image quality rather than true variability. Conversely, low LPIPS may indicate mode collapse or data leakage rather than accurate diversity. Therefore, interpreting LPIPS jointly with FID is essential. SymmFlow achieves the best trade-off between image quality and diversity, reflecting strong generative capability without sacrificing semantic alignment.

\begin{figure*}[!ht]
    \centering
    \includegraphics[width=\linewidth]{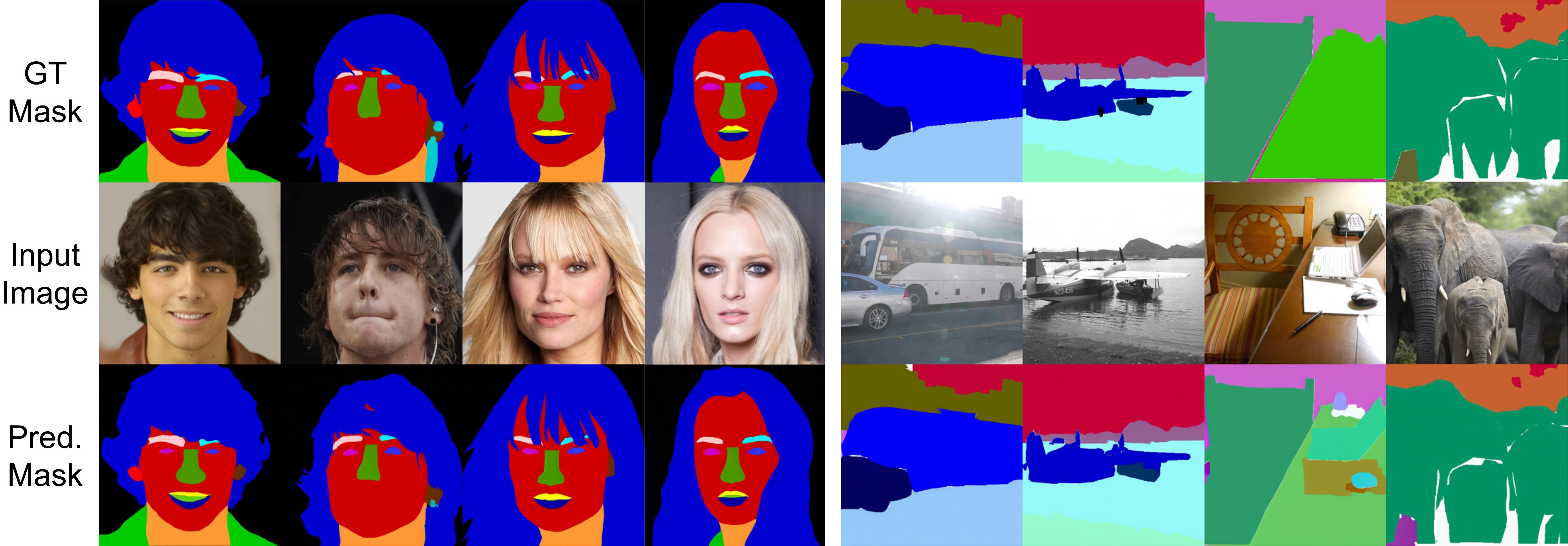}
    \caption{Non-curated segmentation masks generated by the model trained on CelebAMask-HQ~(left) and COCO-stuff~(right). The top row shows the ground-truth segmentation mask. The middle row shows the image used to condition the model. The bottom row shows the segmentations after 25 integration steps with the Euler ODE solver.}
    \label{fig:coco_samples}
\end{figure*}

In the segmentation task, SymmFlow achieves competitive performance compared to specialized segmentation baselines, particularly on COCO-Stuff. As shown in Figure~\ref{fig:coco_samples}, the model demonstrates semantic understanding beyond the ground-truth annotations; for instance, correctly identifying a laptop absent from the label map. However, performance is limited by the low-resolution latent representation (64$\times$64$\times$4), which hampers fine-grained accuracy. This is especially evident for small-area classes such as earrings or partially occluded features like eyebrows and ears, where segmentation quality deteriorates due to insufficient spatial detail. Despite this, the global segmentation structure remains coherent. SemFlow, while also achieving strong performance on the segmentation task, falls short on semantic image synthesis. Quantitative results indicate limited visual fidelity, underscoring the limitations of the architecture in jointly modeling segmentation and generation without further design or training adjustments.

SymmFlow achieves this performance using only 25 function evaluations, whereas most diffusion-based methods require hundreds of denoising steps, resulting in significantly improved inference efficiency, as shown in Appendix~\ref{sec:latency_sup}.

\subsection{Classification}

Table~\ref{tab:class_results} presents the classification performance comparison. With a single inference step, SymmFlow achieves accuracy comparable to the Diffusion Classifier while being significantly more efficient. By increasing the number of steps to just 25, which is 100 times fewer than required by the Diffusion Classifier, SymmFlow convincingly outperforms it on CIFAR-10. This highlights the efficiency of the approach, reducing inference time without sacrificing accuracy. Additionally, the model currently uses a simple conditioning strategy based on grayscale intensities, suggesting further improvements can be achieved with more advanced conditioning mechanisms. In Appendix~\ref{sec:results_sup}, Figures~\ref{fig:mnist} and~\ref{fig:cifar10} present non-curated samples generated by SymmFlow on MNIST and CIFAR-10, illustrating the model’s ability to produce diverse and visually coherent outputs. These findings indicate that SymmFlow continues to be a robust image generator, enabling efficient conditional control.

\begin{table}[!ht]
    \centering
    {\resizebox{\linewidth}{!}{
    \begin{tabular}{l|ccc}
    \toprule
    \textbf{Method} & \textbf{Steps} & \textbf{MNIST} & \textbf{CIFAR-10} \\
    \midrule
    Diffusion Classifier~\cite{li2023your} & 2,750 & --- & 88.5  \\
    \midrule
    \multirow{2}{*}{SymmFlow~(Proposed)} & 1 & 99.3 & 88.2 \\
    & 25 & \textbf{99.6} & \textbf{90.6}\\
    \bottomrule
    \end{tabular}}}
    \caption{Comparison between Diffusion Classifier and SymmFlow on image classification tasks.}
    \label{tab:class_results}
\end{table}

\subsection{Evaluation of Inference Steps}

Table~\ref{tab:impact_gen} evaluates the effect of the number of inference steps on semantic image generation performance across both CelebAMask-HQ and COCO-Stuff. On both datasets, FID and LPIPS scores steadily decrease as the number of steps increases, indicating improvements in image fidelity and structural coherence. In both datasets, the LPIPS trend is particularly revealing: at low step counts, the high LPIPS values do not primarily reflect meaningful diversity, but rather poor visual quality and weak alignment with the conditioning masks. As the number of steps increases, the model produces more coherent and semantically accurate outputs, leading to a decrease in LPIPS. Importantly, the final LPIPS values remain relatively high, but this now reflects genuine variability across samples rather than noise or misalignment, indicating that the model preserves diversity while improving structural fidelity. Appendix~\ref{sec:dequant_sup} analyzes the impact of the dequantization factors, while Appendix~\ref{sec:results_sup} illustrates the qualitative effects of different numbers of sampling steps.

\begin{table}[!ht]
\centering
\resizebox{\linewidth}{!}{
\begin{tabular}{ll|cccccc}
\toprule
\textbf{Dataset} & \textbf{Metric} & \textbf{1} & \textbf{2} & \textbf{5} & \textbf{10} & \textbf{20} & \textbf{25} \\
\midrule
\multirow{2}{*}{CelebA} 
& FID$\downarrow$  & 88.5 & 73.2 & 49.5 & 28.2 & 14.1 & \textbf{11.9} \\
& LPIPS$\downarrow$ & 0.598 & 0.572 & 0.522 & 0.486 & 0.466 & \textbf{0.464} \\
\midrule
\multirow{2}{*}{COCO} 
& FID$\downarrow$ & 102.6 & 83.6 & 44.3 & 18.2 & 8.1  & \textbf{7.0} \\
& LPIPS$\downarrow$ & 0.777 & 0.758 & 0.704 & 0.652 & 0.616 & \textbf{0.609} \\
\bottomrule
\end{tabular}
}
\caption{Semantic image generation performance for different numbers of steps on CelebAMask-HQ and COCO-Stuff.}
\label{tab:impact_gen}
\end{table}

Table~\ref{tab:impact_seg} shows that one-step segmentation on CelebAMask-HQ already yields a respectable 65.3 mIoU, rising to its maximum of 70.3 mIoU by two steps before plateauing. On COCO-Stuff, segmentation quality reaches a solid 38.1 mIoU by five steps and continues to improve modestly, peaking at 40.1 mIoU by twenty steps. While image generation clearly benefits from additional inference steps, these results suggest that, when the sole objective is semantic segmentation, far fewer steps can be employed without sacrificing much accuracy.

\begin{table}[!ht]
\centering
\resizebox{\linewidth}{!}{
\begin{tabular}{ll|cccccc}
\toprule
\textbf{Dataset} & \textbf{Metric} & \textbf{1} & \textbf{2} & \textbf{5} & \textbf{10} & \textbf{20} & \textbf{25} \\
\midrule
\multirow{1}{*}{CelebA} 
& mIoU~$\uparrow$   & 65.3 & \textbf{70.3} & \textbf{70.3} & 69.8 & 69.4 & 69.3 \\
\multirow{1}{*}{COCO} 
& mIoU~$\uparrow$  & 29.3 & 33.8 & 38.1 & 38.9 & \textbf{40.1} & 39.6 \\
\bottomrule
\end{tabular}
}
\caption{Semantic segmentation performance at different numbers of steps on CelebAMask-HQ and COCO-Stuff.}
\label{tab:impact_seg}
\end{table}

The impact of the inference steps in classification is covered in Appendix~\ref{sec:impact_sup}, and summarized in Table~\ref{tab:impact_class}. 

\section{Limitations \& Future Work}
\label{sec:limit}

One limitation of the current approach is the overall model size. Although SymmFlow uses far fewer inference steps than typical diffusion models, it relies on a large pre-trained Stable Diffusion U-Net backbone. Reducing the computational burden by distilling the model into a one-step variant is a natural next step. 
Moreover, a one-step formulation could facilitate the integration of more expressive conditioning mechanisms, enabling effective text-based control and the use of more powerful architectures such as MMDiT~\cite{esser2024scaling}. Additionally, fine-tuning the VAE decoder to better align with semantic masks could improve segmentation accuracy on fine-grained details, such as small or occluded regions.  

Future work includes extending classification evaluation from the current proof-of-concept stage to datasets such as Food-101~\cite{bossard2014food}, 
ImageNet-1K~\cite{deng2009imagenet}, and ObjectNet~\cite{barbu2019objectnet}, while also refining the semantic label encoding strategy to improve conditioning. Furthermore, evaluating the model performance on depth estimation tasks, similar to DepthFM, would further demonstrate its versatility. Beyond segmentation, the bi-directionality of the model presents opportunities for applications such as image editing.

\section{Conclusions}
\label{sec:conclusions}

This work introduces Symmetrical Flow Matching, a unified framework that models segmentation, classification, and image generation as opposing flows within a single architecture. Leveraging a bi-directional formulation, SymmFlow enables efficient semantic reasoning while preserving the flexibility required for high-fidelity generation. Unlike prior approaches that impose rigid one-to-one mappings, it supports various conditioning strategies, including pixel- and image-level supervision. Experimental results show that SymmFlow achieves state-of-the-art performance in semantic image synthesis with only 25~inference steps, competitive segmentation accuracy despite operating in a low-resolution latent space, and promising classification results. These results show that flow-based generative models can simultaneously support generation and discrimination within a unified framework. Future work will further explore its classification potential and extend the framework to problems such as depth estimation and semantic image editing.

\section*{Acknowledgments}

This work used the Dutch national e-infrastructure with the support of the SURF Cooperative using grant no. EINF-13640. This research was funded by the European Xecs Eureka TASTI Project.



\setcounter{page}{1}

\setcounter{secnumdepth}{2}
\appendix
\section*{Appendix/supplemental material}
\label{sec:appendix}

The supplementary material is organized as follows: Appendix~\ref{sec:data_sup} covers the datasets used in this work. Appendix~\ref{sec:implementation_sup} describes the implementation details of the employed models and the computational resources required for training and evaluating the models. Appendix~\ref{sec:latency_sup} offers details on model image generation latency. Appendix~\ref {sec:metrics_sup} provides additional information about how the semantic image synthesis metrics are computed. Appendix~\ref{sec:dequant_sup} analyses the impact of the dequantization term $\beta$ in the image generation performance. Appendix~\ref{sec:impact_sup} evaluates the impact of inference steps on classification performance. Appendix~\ref{sec:results_sup} contains additional qualitative results of sampling.

\section{Datasets}
\label{sec:data_sup}

This section provides details on the datasets used for classification and semantic segmentation and generation. For classification, we use MNIST~\cite{deng2012mnist} and CIFAR-10~\cite{krizhevsky2010cifar}, while for segmentation, we evaluate on CelebAMask-HQ~\cite{lee2020maskgan} and COCO-Stuff~\cite{caesar2018coco}. The datasets are summarized in Table~\ref{tab:datasets}.

\begin{table}[h]
    \centering
    {\resizebox{\linewidth}{!}{
    \begin{tabular}{l|ccc}
        \toprule    
        \textbf{Dataset} & \textbf{Train Images} & \textbf{Test Images} & \textbf{Classes} \\
        \midrule
        MNIST & 60,000 & 10,000 & 10 \\
        CIFAR-10 & 50,000 & 10,000 & 10 \\
        \hline
        CelebAMask-HQ & 24,183 & 2,824 & 19 \\
        COCO-stuff & 118,287 & 5,000 & 171 \\
        \bottomrule
    \end{tabular}}}
    \caption{Summary of the benchmark datasets.}
    \label{tab:datasets}
\end{table}

The classification datasets, MNIST and CIFAR-10, consist of low-resolution images (32$\times$32), which provide a controlled setting for evaluating fundamental classification capabilities. In contrast, the segmentation datasets, CelebAMask-HQ and COCO-Stuff, contain higher-resolution images (512$\times$512), allowing for a more detailed assessment of semantic segmentation performance.

COCO-Stuff does not provide RGB masks, so a custom color palette is generated, assigning a unique color to each of the 171 classes. This enables visualization and evaluation of segmentation predictions in a manner consistent with other datasets that provide pre-defined RGB masks.

MNIST and CIFAR-10 were automatically downloaded using the PyTorch dataloader provided in the \texttt{Torchvision} library. CelebAMask-HQ was obtained from Hugging Face\footnote{\url{https://huggingface.co/datasets/eurecom-ds/celeba_hq_mask}}, while COCO-Stuff was downloaded from the official dataset repository\footnote{\url{https://github.com/nightrome/cocostuff}}.

\section{Implementation Details}
\label{sec:implementation_sup}

The MNIST model was trained on a system equipped with an NVIDIA RTX 2080 Ti GPU with 11GB of VRAM, an Intel Xeon Silver 4216 CPU (2.10 GHz), and 192GB of RAM. The CIFAR-10 model was trained on a system featuring an NVIDIA A100-SXM4 GPU with 40GB of VRAM, an Intel Xeon Platinum 8360Y CPU, and 512GB of RAM. The CelebAMask-HQ and COCO-stuff models were trained on a system with four NVIDIA H100 GPUs, each with 94GB of VRAM, an AMD EPYC 9334 CPU, and 768GB of RAM.

For pixel-space implementations on MNIST and CIFAR-10, we use the U-Net architecture proposed by Dhariwal~\emph{et al.}~\cite{dhariwal2021diffusion}, which has been widely adopted for diffusion-based generative modeling. The architecture consists of a series of residual blocks, self-attention layers, and group normalization, enabling effective denoising and feature extraction across multiple resolutions. Table~\ref{tab:hyperparams} presents the hyperparameters used for training these two models. To ensure reproducibility, we will also make the code publicly available.

\begin{table}[!t]
    \centering
    {\resizebox{\linewidth}{!}{
    \begin{tabular}{lcc}
        \toprule
        \textbf{Hyperparameter} & \textbf{MNIST} & \textbf{CIFAR-10} \\
        \midrule
        Channels & 32 & 256 \\
        Depth & 2 & 2 \\
        Channels Multiple & 1,2,2,2 & 1,2,2,2 \\
        Heads & 4 & 4 \\
        Head Channels & 64 & 64 \\
        Attention Resolution & 16 & 16 \\
        Dropout & 0.0 & 0.0 \\
        Batch Size & 512 & 256 \\
        GPUs & 1 & 1 \\
        Epochs & 1000 & 200 \\
        Learning Rate & 5e-4 & 3e-4  \\
        Learning Rate Scheduler & Cosine Annealing & Cosine Annealing  \\
        Warmup Epochs & 100 & 100  \\
        $\beta$ & 4 & 4 \\
        \bottomrule
    \end{tabular}}}
    \caption{Hyperparameters used for training each model.}
    \label{tab:hyperparams}
\end{table}

For latent-space implementations, we use the pre-trained Variational Autoencoder~(VAE) from Stable Diffusion, which efficiently compresses high-dimensional image data into a lower-dimensional latent space while preserving perceptual quality. The VAE can be downloaded from HuggingFace~\footnote{\url{https://huggingface.co/stabilityai/sd-vae-ft-mse}}. The U-Net and pretrained weights correspond to those used in Stable Diffusion 2.1 and are available on HuggingFace~\footnote{\url{https://huggingface.co/stabilityai/stable-diffusion-2-1}}. After loading the weights, the number of input channels in the first layer and the number of output channels in the last layer are doubled. These models were trained with mixed precision while supporting multi-GPU training. This was made possible through the \texttt{Accelerate} library. Table~\ref{tab:hyperparams_sd} presents the training hyperparameters for both models.

\begin{table}[!ht]
    \centering
    {\resizebox{\linewidth}{!}{
    \begin{tabular}{lcc}
        \toprule
        \textbf{Hyperparameter} & \textbf{CelebAMask-HQ} & \textbf{COCO-stuff} \\
        \midrule
        Batch Size & 32 & 32 \\
        GPUs & 2 & 4 \\
        Epochs & 200 & 200 \\
        Learning Rate & 8e-5 & 8e-5  \\
        Learning Rate Scheduler & Cosine Annealing & Cosine Annealing  \\
        Warmup Epochs & 10 & 10 \\
        $\beta$ & 10 & 6 \\
        \bottomrule
    \end{tabular}}}
    \caption{Hyperparameters used for training each model.}
    \label{tab:hyperparams_sd}
\end{table}

\section{Image Generation Latency}
\label{sec:latency_sup}

This experiment measures the average time required to generate a single image, including both the forward pass through the VAE and the generator. Latency was evaluated for ControlNet, SemFlow, and SymmFlow, as these models achieve competitive performance with a practical number of inference steps. Other baselines were excluded from this comparison due to either excessively high computational cost or insufficient generation quality, as observed in the one-step variants. The reported results reflect the mean generation time per image under identical hardware conditions, on a system featuring an NVIDIA A100-SXM4 GPU with 40GB of VRAM, an Intel Xeon Platinum 8360Y CPU, and 512GB of RAM.

The results in Table~\ref{tab:lat} show that SymmFlow achieves a substantial reduction in generation time compared to ControlNet. While SymmFlow and SemFlow exhibit nearly identical inference times, the former delivers significantly higher visual quality and consistency.

\begin{table}[!ht]
    \centering
    {\resizebox{\linewidth}{!}{
    \begin{tabular}{l|c c}
        \toprule
        \textbf{Model} & \textbf{Latency~(ms)} \\
        \midrule
        ControlNet~\cite{zhang2023adding} & 5,483.8 \\
        SemFlow~\cite{wang2024semflow} & 2,140.1 \\
        SymmFlow~(Proposed) & 2,156.9 \\
        \bottomrule
    \end{tabular}}}
    \caption{Image generation latency, measured as the average time required to generate a single image.}
    \label{tab:lat}
\end{table}

\section{Metrics}
\label{sec:metrics_sup}

For the quantitative evaluation of semantic image synthesis, 50,000 images are generated from the available validation masks, which are used to compute perceptual metrics. The FID is evaluated by comparing the generated samples with the real images from the dataset, assessing overall visual realism and distribution alignment. The LPIPS metric is computed against the validation set by comparing each synthesized image to its corresponding ground-truth image derived from the same mask. For the CIFAR-10 dataset, 50,000 images are sampled instead, evenly distributed across the ten classes~(5,000 per class) to ensure balanced evaluation.

\section{Dequantization Impact on Generation}
\label{sec:dequant_sup}

This experiment evaluates the impact of different dequantization factors $\beta$ on semantic image generation performance for the CelebAMask-HQ and COCO-Stuff datasets. CelebAMask-HQ was trained with a dequantization factor of $\beta = 10.0$, while COCO-Stuff used $\beta = 6.0$. The results in Table~\ref{tab:impact_deq} demonstrate that a moderate increase in $\beta$ at inference time can yield improved generation quality, as reflected by lower FID values. This observation suggests that adjusting the dequantization strength can help balance sample diversity and reconstruction fidelity. A more systematic investigation of $\beta$ during training is necessary to better characterize its influence on the generative process and the resulting image quality.

\begin{table}[!ht]
\centering
\resizebox{\linewidth}{!}{
\begin{tabular}{ll|ccccc}
\toprule
\textbf{Dataset} & \textbf{Metric} & \textbf{4.0} & \textbf{6.0} & \textbf{8.0} & \textbf{10.0} & \textbf{12.0} \\
\midrule
\multirow{2}{*}{CelebA} 
& FID$\downarrow$  & 22.6  & 22.1  & 17.0 & 11.9 & \textbf{10.1} \\
& LPIPS$\uparrow$ & 0.477 & 0.471  & \textbf{0.464}  & \textbf{0.464} & 0.471 \\
\midrule
\multirow{2}{*}{COCO} 
& FID$\downarrow$ & 10.6 & 7.0 & \textbf{6.2} & 6.8 & 8.0 \\
& LPIPS$\uparrow$ & 0.627 & 0.609 & \textbf{0.606} & 0.611 & 0.619\\
\bottomrule
\end{tabular}
}
\caption{Semantic image generation performance for different values of $\beta$ on CelebAMask-HQ and COCO-Stuff. These results were obtained after 25 integration steps with the Euler ODE solver.}
\label{tab:impact_deq}
\end{table}

\section{Inference Steps on Classification}
\label{sec:impact_sup}

Table~\ref{tab:impact_class} evaluates the effect of inference steps on classification accuracy for the MNIST and CIFAR-10 data. On CIFAR-10, increasing the number of steps initially degrades performance. This is consistent with observations in the Toy Dataset experiment, where additional steps led to misalignment between the reverse flow and the correct decision regions. However, as more steps are introduced, performance begins to recover. This suggests that the steps closest to the original data distribution \(X\) play a crucial role in guiding the reverse process, and when too few steps are used, the model fails to properly leverage this information. When more steps are introduced, the flow receives a stronger signal from these informative regions, allowing it to better reconstruct the class structure.

\begin{table}[!ht]
    \centering
    {\resizebox{\linewidth}{!}{
    \begin{tabular}{ll|c c c c c c}
        \toprule
        Dataset & Metric & 1 & 2 & 5 & 10 & 20 & 25 \\
        \midrule
        MNIST & Acc.~$\uparrow$ & 99.3 & 99.4  & 99.5  & 99.4 & 99.5 & \textbf{99.6}  \\
        CIFAR-10 & Acc.~$\uparrow$ & 88.2 & 52.3  & 63.5  & 74.9 & 89.4 & \textbf{90.6}  \\
        \bottomrule
    \end{tabular}}}
    \caption{Classification accuracy, measured for different numbers of steps on MNIST and CIFAR-10.}
    \label{tab:impact_class}
\end{table}

For the MNIST data, this effect is less pronounced. Although evolving towards a Gaussian distribution causes class boundaries to blur, the simplicity of digit shapes allows for correct classification across different step counts. These results suggest that, although single-step inference can be effective, datasets with more complex semantics benefit from a sufficient number of steps to preserve meaningful class information.

\section{Additional Qualitative Results}
\label{sec:results_sup}

\begin{figure*}
    \centering
    \includegraphics[width=0.75\linewidth]{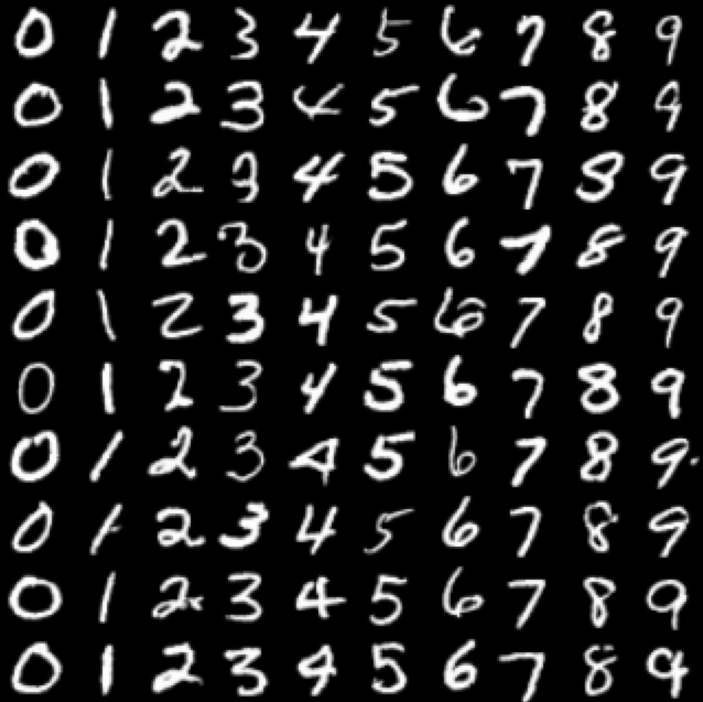}
    \caption{Non-curated samples of the SymmFlow model trained on MNIST.}
    \label{fig:mnist}
\end{figure*}

\begin{figure*}
    \centering
    \includegraphics[width=0.75\linewidth]{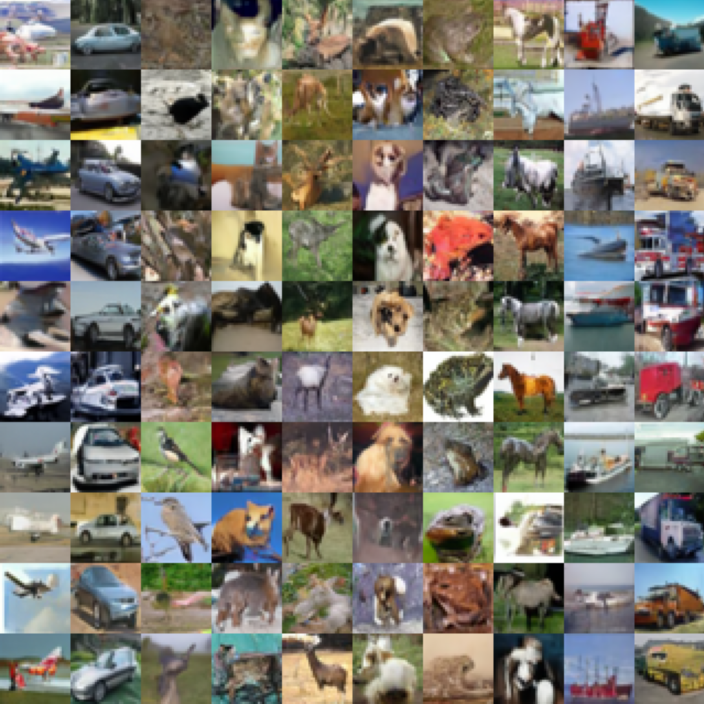}
    \caption{Non-curated samples of the SymmFlow model trained on CIFAR-10.}
    \label{fig:cifar10}
\end{figure*}

\begin{figure*}
    \centering
    \includegraphics[width=0.85\linewidth]{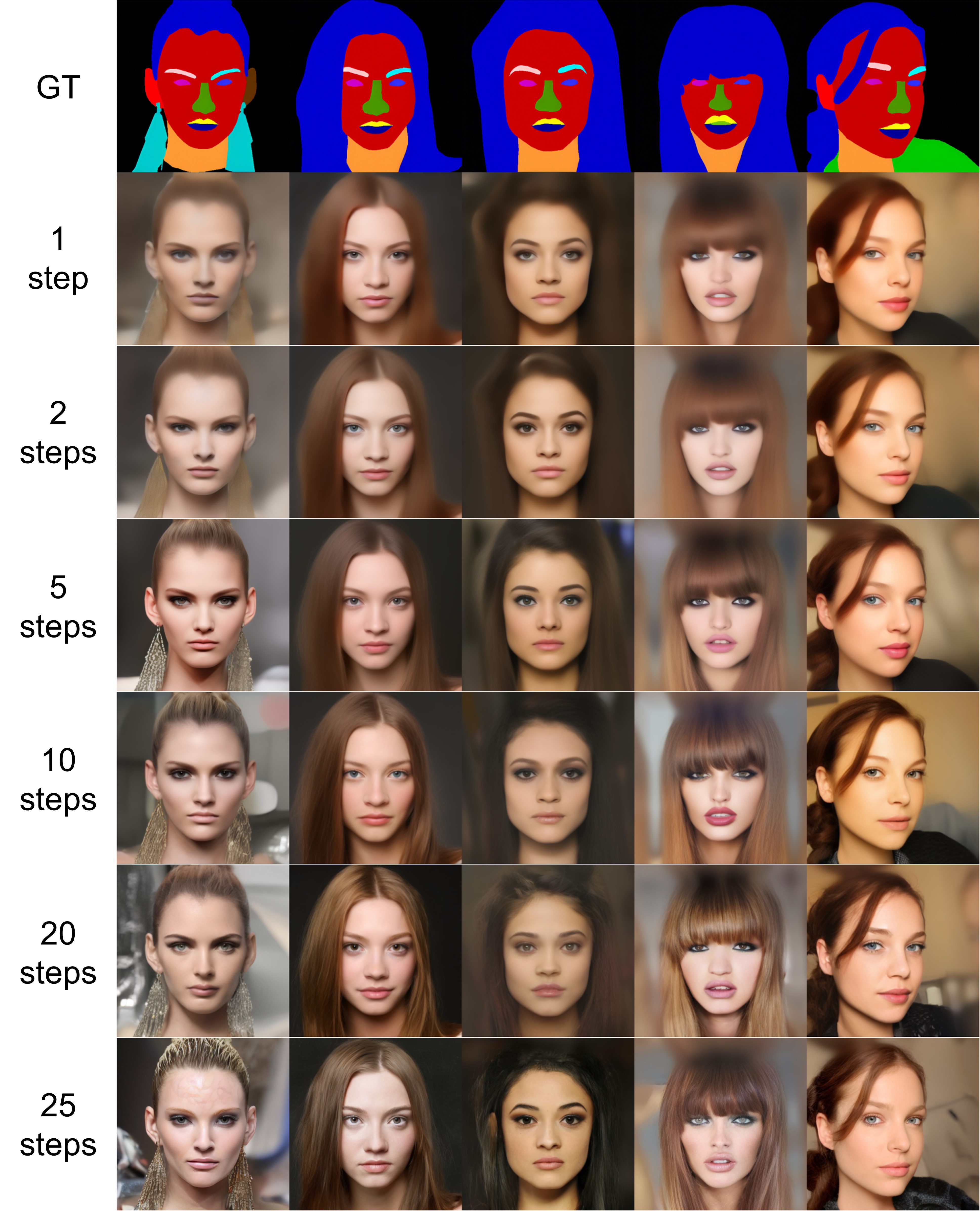}
    \caption{Non-curated samples of the SymmFlow model trained on CelebAMask-HQ using different sampling steps.}
    \label{fig:steps_celeb}
\end{figure*}

\begin{figure*}
    \centering
    \includegraphics[width=0.85\linewidth]{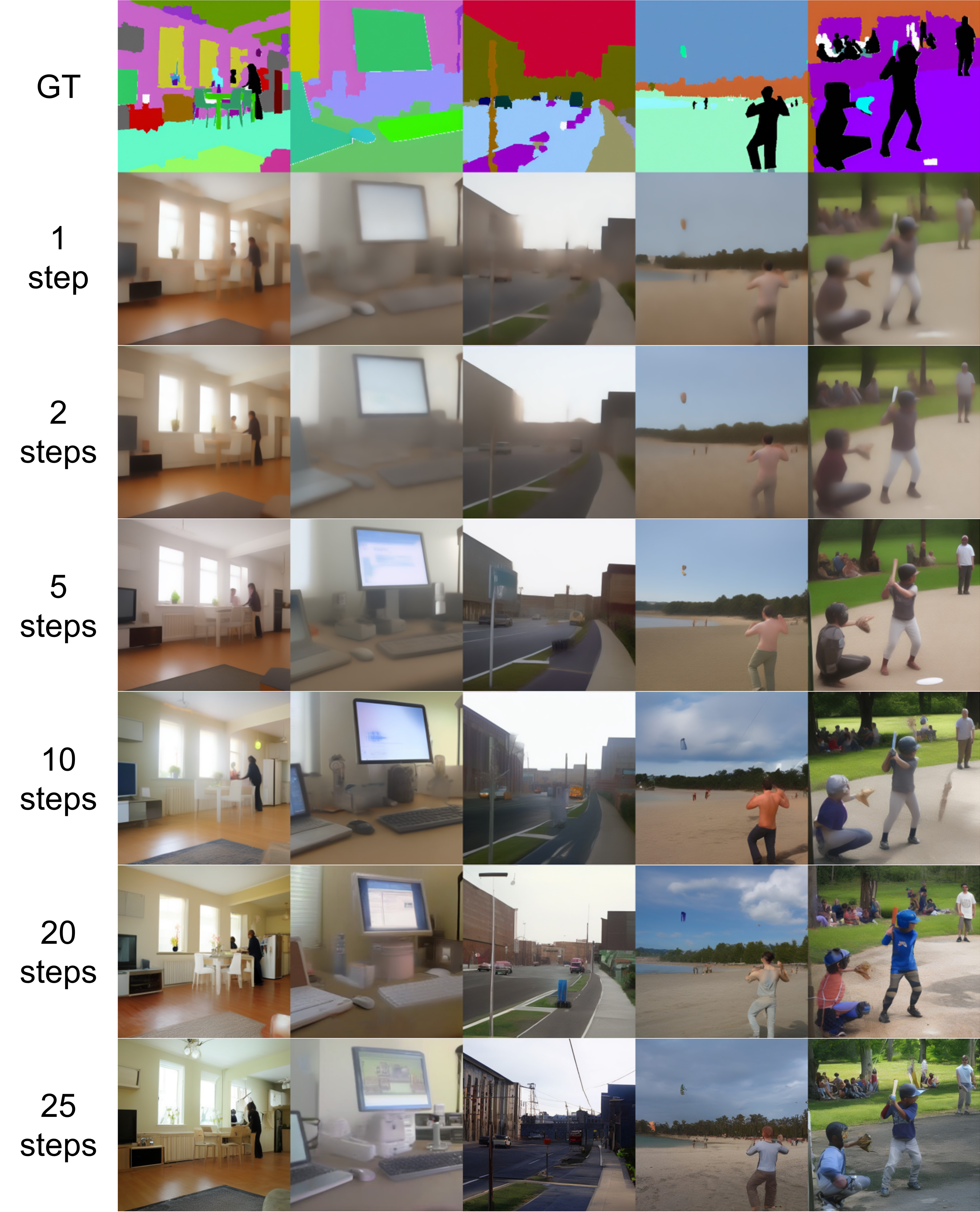}
    \caption{Non-curated samples of the SymmFlow model trained on COCO-stuff using different sampling steps.}
    \label{fig:steps_coco}
\end{figure*}

\begin{figure*}
    \centering
    \includegraphics[width=0.85\linewidth]{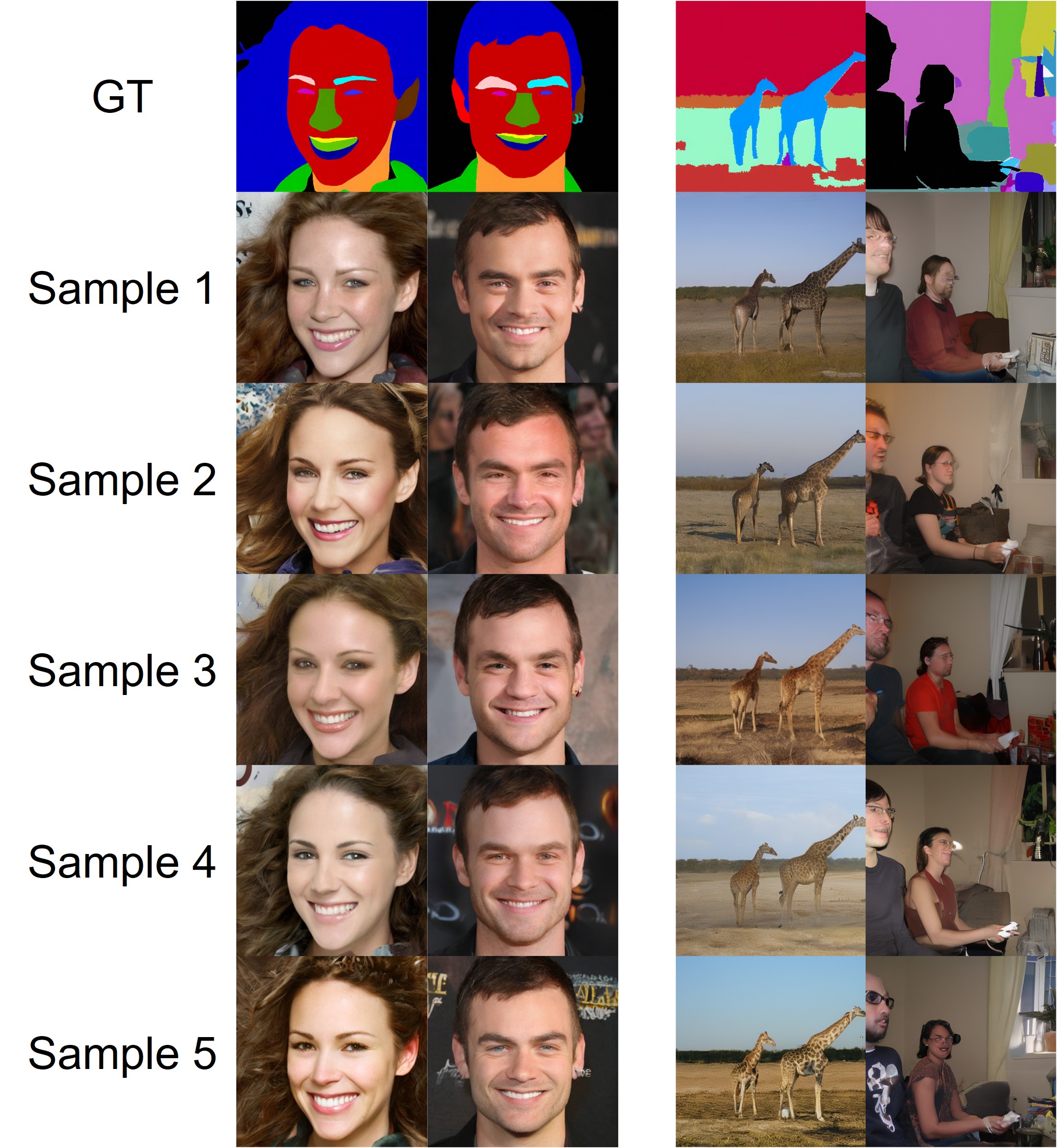}
    \caption{Non-curated images generated by the model trained on CelebAMask-HQ~(left) and COCO-stuff~(right). The top row shows the semantic mask used to condition the model. The other rows show the generated samples after 25 integration steps with the Euler ODE solver.}
    \label{fig:diversity}
\end{figure*}

Figure~\ref{fig:mnist} contains conditional samples generated by the model trained on MNIST. Figure~\ref{fig:cifar10} shows conditional samples from the model trained on CIFAR-10. Figures~\ref{fig:steps_celeb} and~\ref{fig:steps_coco} demonstrate the effect of sampling steps on the sample quality for CelebAMask-HQ and COCO-stuff, respectively. Furthermore, Figure~\ref{fig:diversity} showcases the diversity of the samples produced by SymmFlow, when conditioned on the same semantic mask.

\end{document}